\documentclass[10pt,twocolumn,letterpaper]{article}

\usepackage{cvpr}              

\usepackage[dvipsnames]{xcolor}
\usepackage{multirow}

\definecolor{cvprblue}{rgb}{0.21,0.49,0.74}
\usepackage{hyperref}
\hypersetup{pagebackref,breaklinks,colorlinks,citecolor=cvprblue}

\def\paperID{5473} 
\def\confName{CVPR}
\def\confYear{2024}

\title{ODM: A Text-Image Further Alignment Pre-training Approach for Scene Text Detection and Spotting}

\author{
Chen Duan\thanks{First Author and Second Author contribute equally to this work.}\\
{\tt\small duanchen02@meituan.com}
\and
Pei Fu\footnotemark[1]\\
{\tt\small fupei@meituan.com}
\and
Shan Guo\\
{\tt\small guoshan05@meituan.com}
\and
Qianyi Jiang\\
{\tt\small jiangqianyi02@meituan.com}
\and
Xiaoming Wei\\
{\tt\small weixiaoming@meituan.com}
}

\begin{document}
\maketitle
\begin{abstract}
In recent years, text-image joint pre-training techniques have shown promising results in various tasks. However, in Optical Character Recognition (OCR) tasks, aligning text instances with their corresponding text regions in images poses a challenge, as it requires effective alignment between text and OCR-Text (referring to the text in images as OCR-Text to distinguish from the text in natural language) rather than a holistic understanding of the overall image content. In this paper, we propose a new pre-training method called \textbf{O}CR-Text \textbf{D}estylization \textbf{M}odeling (ODM) that transfers diverse styles of text found in images to a uniform style based on the text prompt. With ODM, we achieve better alignment between text and OCR-Text and enable pre-trained models to adapt to the complex and diverse styles of scene text detection and spotting tasks. Additionally, we have designed a new labeling generation method specifically for ODM and combined it with our proposed Text-Controller module to address the challenge of annotation costs in OCR tasks, allowing a larger amount of unlabeled data to participate in pre-training. Extensive experiments on multiple public datasets demonstrate that our method significantly improves performance and outperforms current pre-training methods in scene text detection and spotting tasks. Code is available at \href{https://github.com/PriNing/ODM}{ODM}.

\end{abstract}    
\section{Introduction}
\label{sec:intro}
Optical Character Recognition (OCR) has garnered significant attention in the field of computer vision for its remarkable performance in automated data entry, document analysis, instant translation, and more. Most existing methods for obtaining OCR results involve a two-stage process, including a text detection model and a text recognition model, or directly utilizing an end-to-end text spotting model. Furthermore, many of these methods initialize the model through pre-training on the ImageNet dataset \cite{imagenet}.

\begin{figure}[t]
  \centering
   \includegraphics[width=1.\linewidth]{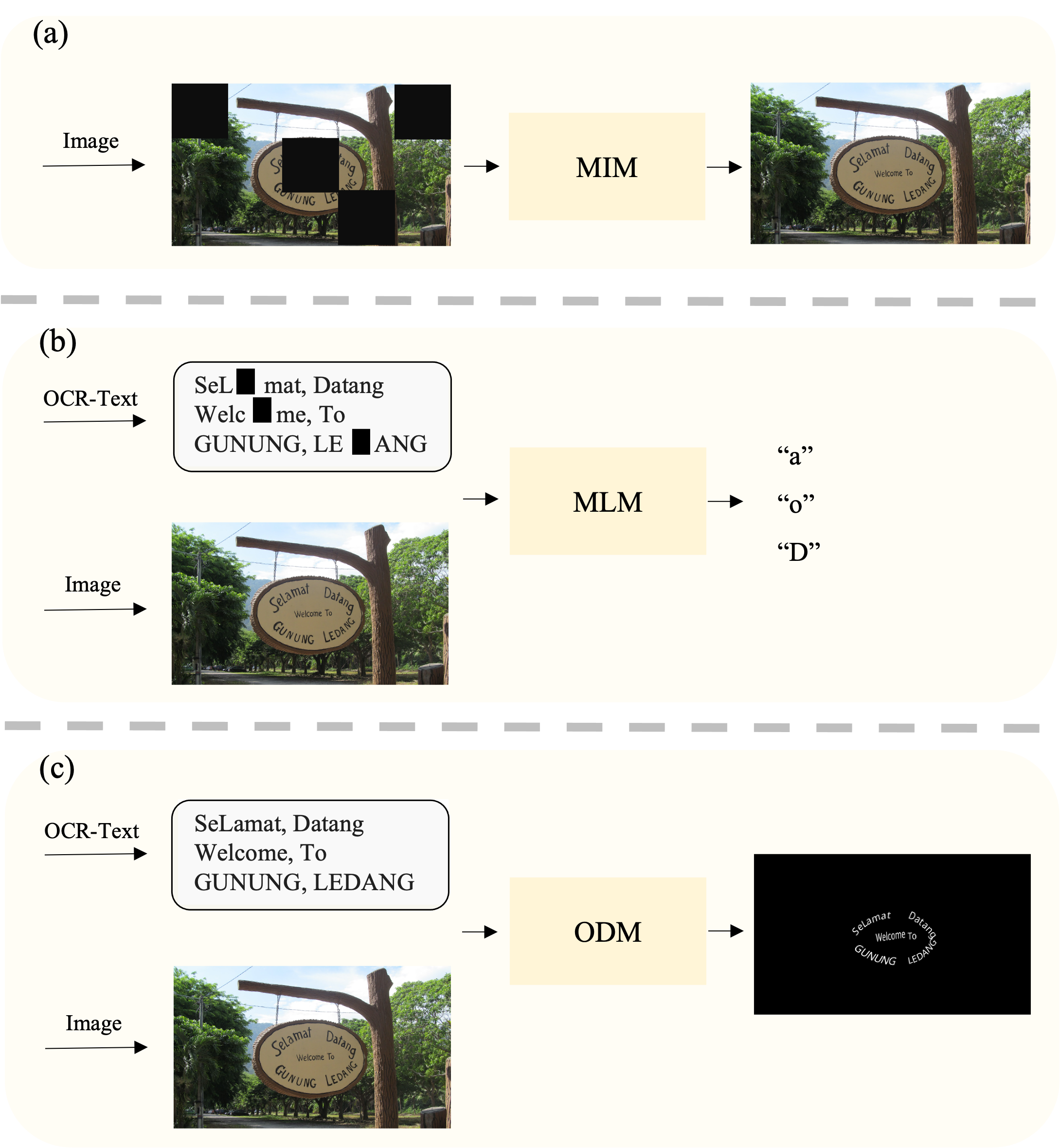}
   \caption{Comparisons of different pre-training strategies. (a) Obtain the pre-trained model through mask image modeling, taking only image embeddings as inputs. (b) Obtain the pre-trained model through mask language modeling, which simultaneously takes both OCR-Text and image as inputs. (c) Our approach obtains the pre-trained model through OCR-Text destylization modeling.}
   \label{fig:onecol}
\end{figure}

Pre-training techniques have recently gained significant attention for their outstanding performance across a wide range of computer vision tasks. Two commonly used pre-training methods in computer vision are: (1) Masked Image Modeling (MIM) pre-training \cite{beit,simmim,mae,efficient}, which focuses on learning visual contextualized representations solely from images. This method is typically applied to vision-dominated tasks, such as image classification. (2) Masked Language Modeling (MLM) \cite{bert,layoutlmv2,language}, which utilizes both text and image as inputs and extracts semantic information from both modalities. However, when applying these pre-training methods in the OCR field, two specific issues arise: (1) With the MIM-based method, there are instances where the text in the image is completely obscured by the masked patch, primarily due to the relatively small proportion of the text. This has the potential to hinder the pre-trained model's ability to effectively learn textual feature information. (2) The MLM-based method, although achieving weakly supervised training by masking the text input, does not explicitly exploit text location information during training. This can lead to ineffective alignment between text and image features, as well as inadequate handling of image information. These challenges highlight the need for innovative approaches that specifically address the unique requirements of OCR.

\begin{figure}[t]
  \centering
  \begin{subfigure}{0.24\linewidth}
    \includegraphics[width=1.\linewidth]{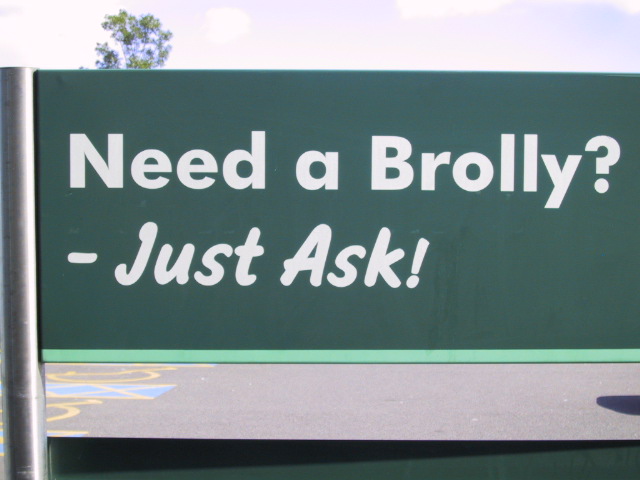}
    \label{fig:short-a}
  \end{subfigure}
  \begin{subfigure}{0.24\linewidth}
    \includegraphics[width=1.\linewidth]{pic/label_gen/1045.jpg}
    \label{fig:short-b}
  \end{subfigure}
  \begin{subfigure}{0.24\linewidth}
    \includegraphics[width=1.\linewidth]{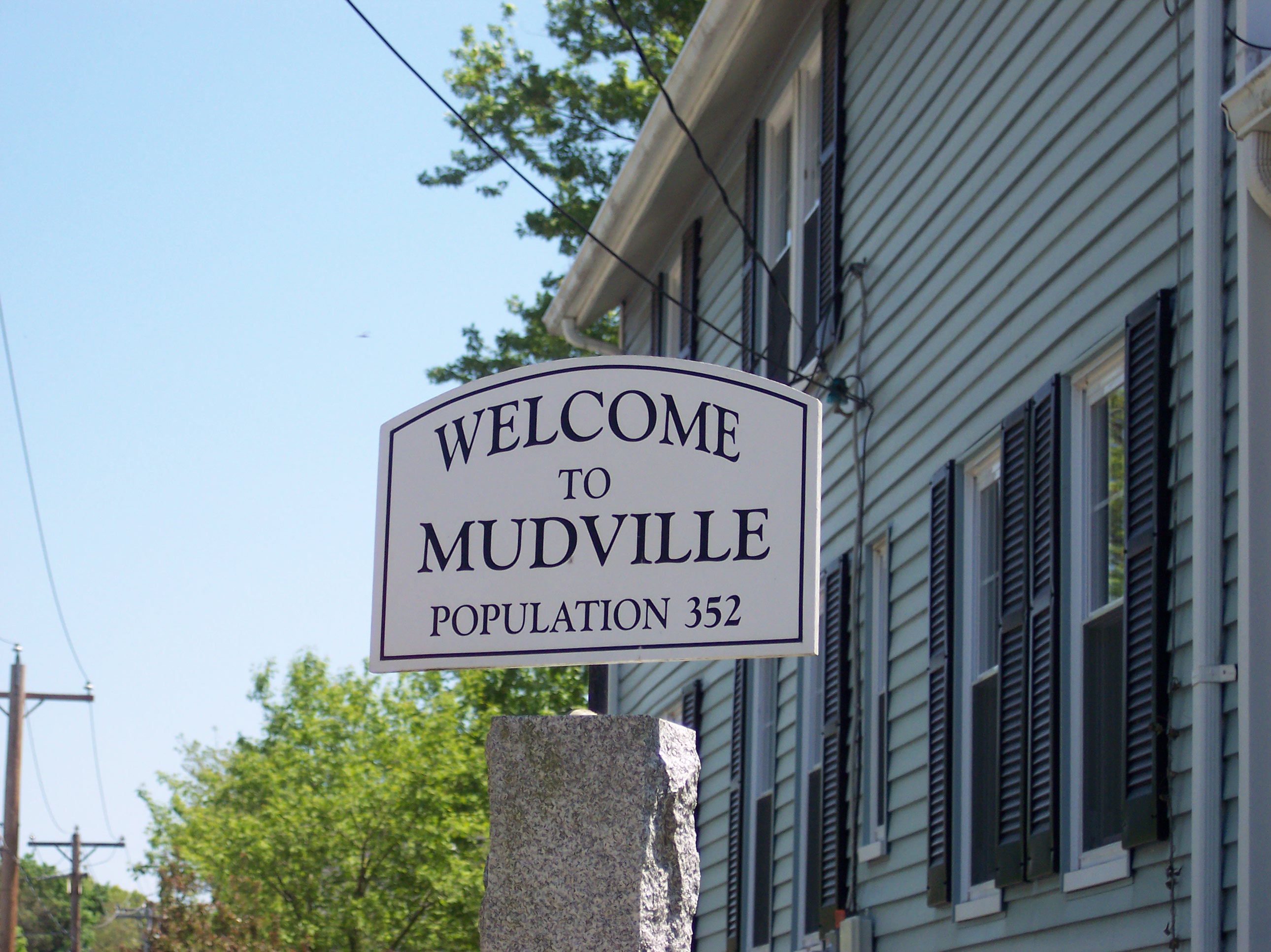}
    \label{fig:short-c}
  \end{subfigure}
  \begin{subfigure}{0.24\linewidth}
    \includegraphics[width=1.\linewidth]{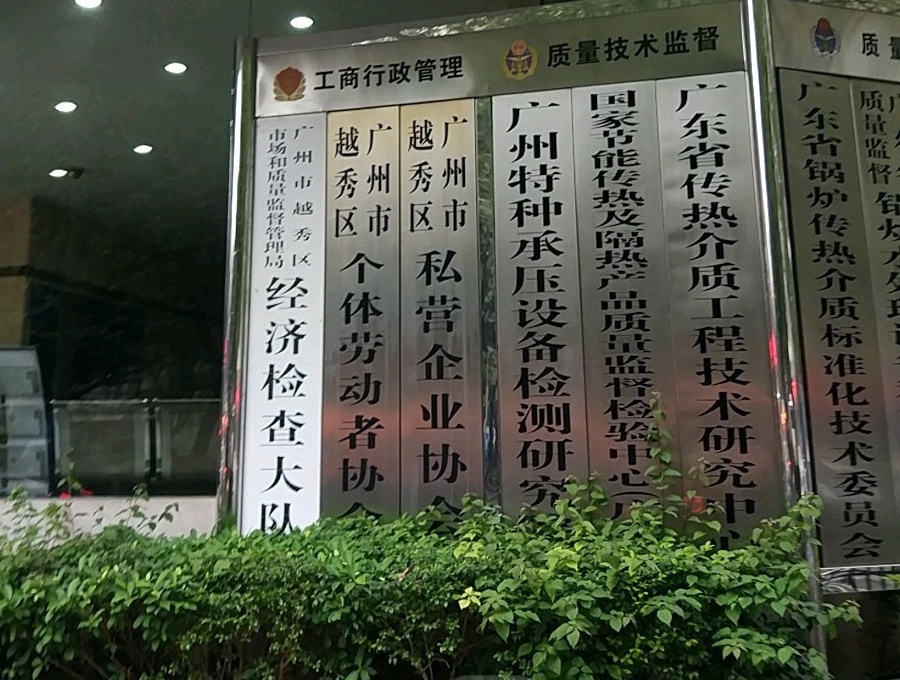}
    \label{fig:short-g}
  \end{subfigure}
  \begin{subfigure}{0.24\linewidth}
    \includegraphics[width=1.\linewidth]{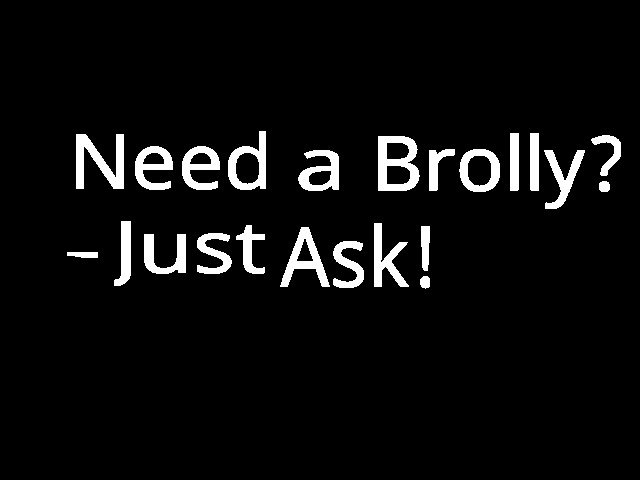}
    \caption{}
    \label{fig:short-d}
  \end{subfigure}
  \begin{subfigure}{0.24\linewidth}
    \includegraphics[width=1.\linewidth]{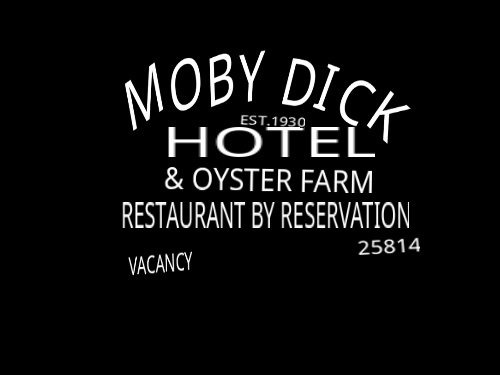}
    \caption{}
    \label{fig:short-e}
  \end{subfigure}
  \begin{subfigure}{0.24\linewidth}
    \includegraphics[width=1.\linewidth]{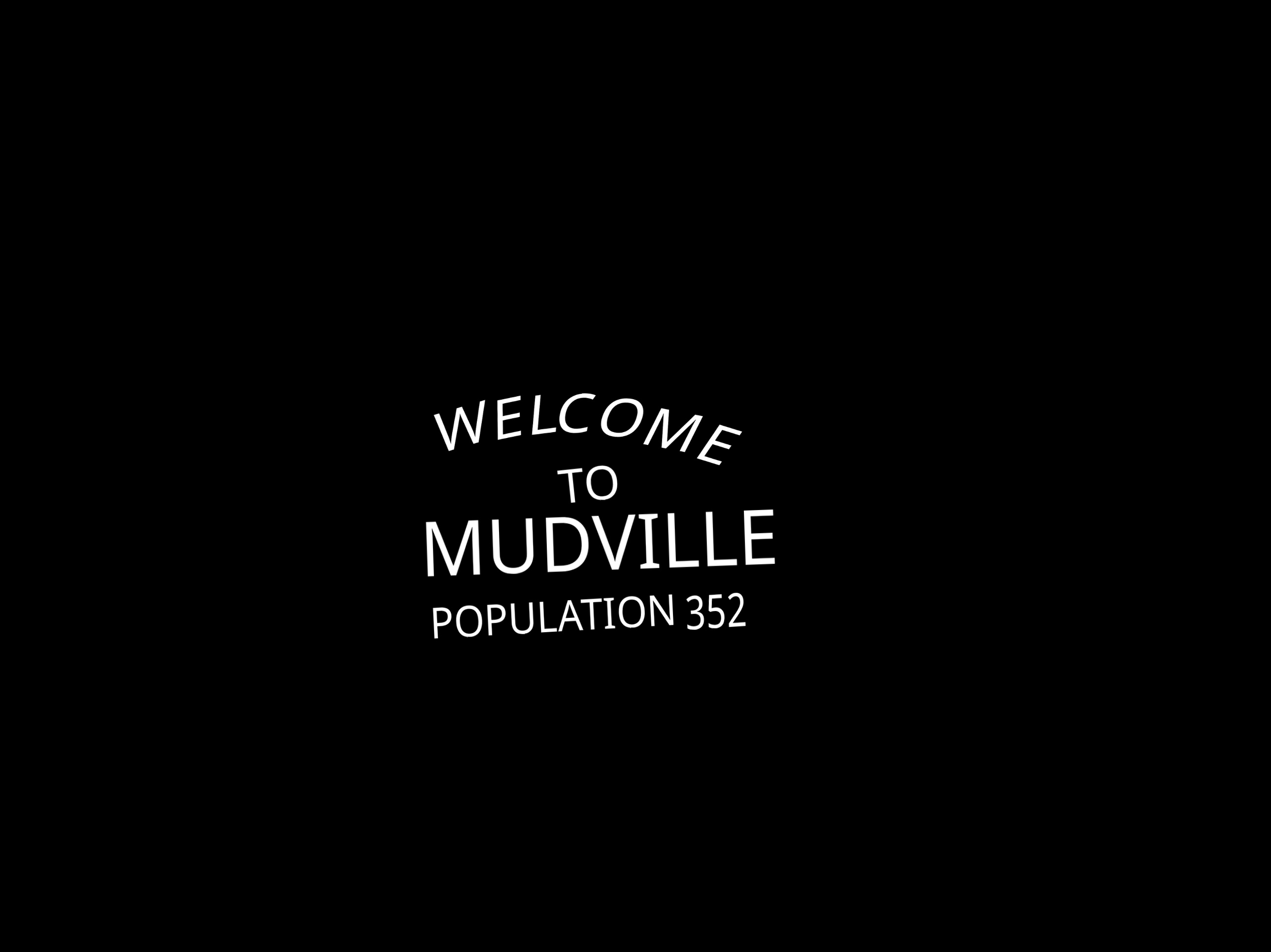}
    \caption{}
    \label{fig:short-f}
  \end{subfigure}
  \begin{subfigure}{0.24\linewidth}
    \includegraphics[width=1.\linewidth]{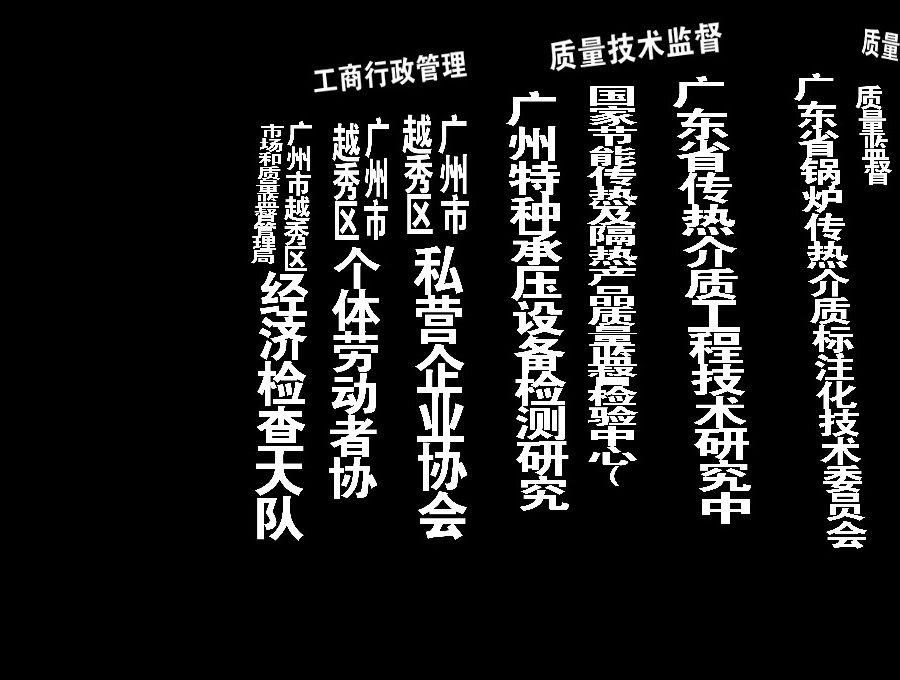}
    \caption{}
    \label{fig:short-h}
  \end{subfigure}
  \caption{The upper row and lower row represent the original images and their corresponding destylized labels, respectively. (a), (b), (c), and (d) are taken from the ICDAR15 \cite{karatzas2015icdar}, CTW1500 \cite{liu2019curved}, TotalText \cite{ch2020total}, and LSVT \cite{sun2019icdar} datasets, respectively.}
  \label{fig:label_gen}
\end{figure}

In this paper, we propose a novel pre-training technique called OCR-Text Destylization Model (ODM) to address the challenges of text-image alignment in OCR tasks. As illustrated in \cref{fig:onecol}, unlike existing MIM and MLM methods, ODM introduces a new pixel-level image reconstruction modeling based on text prompts. Since OCR tasks primarily focus on the text within the image while considering other pixels irrelevant, ODM aims to reconstruct a binary image that removes the text style and enforces alignment between the text and OCR-Text. This is achieved by utilizing a pixel-level reconstruction approach instead of the traditional three-channel reconstruction. To further enhance the model's understanding of the text, we propose a Text-Controller module. This module guides the image encoder to identify and interpret the OCR-Text, facilitating the alignment between the text and OCR-Text. Additionally, we have designed a novel method for generating ODM labels, effectively addressing the issue of inadequate pixel-level labels in the dataset. By leveraging font files, text, and location labels, we generate binary images with a unified font style, as illustrated in \cref{fig:label_gen}, which showcases some examples of OCR-Text destylization images. These advancements in ODM and the Text-Controller module, along with the novel label generation method, contribute to improved text-image alignment in OCR tasks.

In summary, the main contributions are three-fold:

(1) We propose a simple yet effective pre-training method called ODM, which focuses on learning features specifically for OCR-Text. By using pixel-level labels with a uniform style, we successfully destylize OCR-Text, improving text comprehension. This crucial feature information enables the pre-trained model to adapt well to various scenarios in text detection and spotting tasks.

(2) We introduce a novel Text-Controller module that helps regulate the model's output, enhancing its understanding of OCR-Text. With this module, our method does not require a perfect match between the input image and the text pair. As a result, we can utilize weakly annotated data (i.e., using other OCR recognition engines to obtain the text and location in the image and filter it based on recognition confidence, text size, etc.), which can greatly reduce the annotation cost.

(3) Experimental results on public datasets demonstrate that ODM delivers outstanding performance and surpasses existing pre-training techniques across a range of scene text detection and spotting datasets.

\section{Related Work}
\label{sec:RelatedWork}

\begin{figure*}[htp]
  \centering
  \includegraphics[width=1\linewidth]{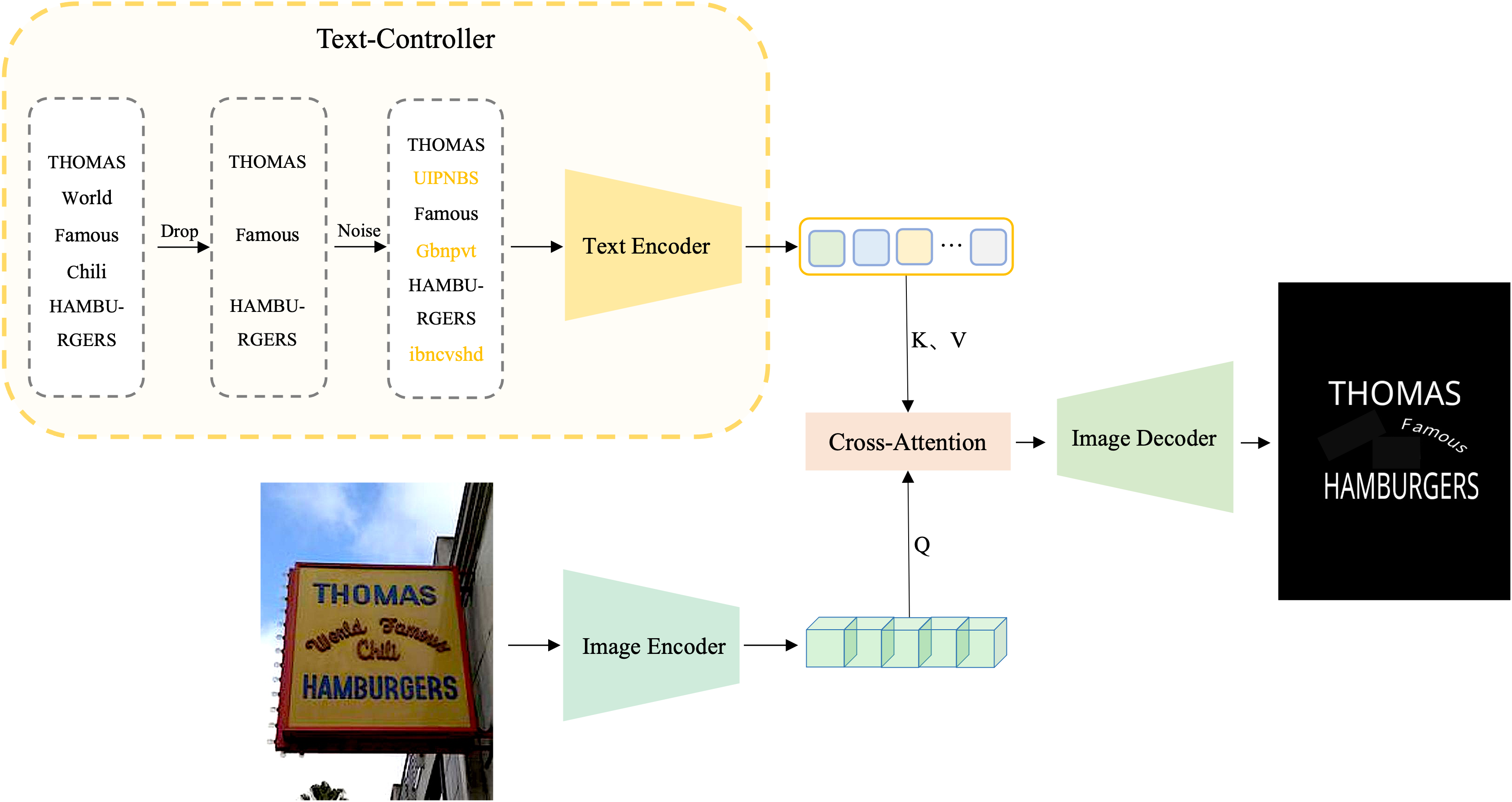}
  \caption{The overall architecture of ODM. The text is encoded by the Text-Controller to get the encoded text features, and the image is encoded by the image encoder to get the encoded image features. The text features and image features interact through cross-attention, and finally output destylization binary image.}
  \label{fig:pipeline}
\end{figure*}


\noindent\textbf{Scene Text Detection.} Scene text detectors based on deep learning can primarily be categorized into regression based 
\cite{2017detecting,2017textboxes,2017east,2018arbitrary,2021most,2022few} and segmentation based \cite{2019textfield,2019pse,2019pan,2020db,2021fce,2022dbnetpp,2023tcm} methods. Regression-based methods perceive scene text as an object and directly regress the bounding box of the text instance. Segmentation-based methods treat the text detection task as a semantic segmentation problem. They obtain a segmentation map by directly segmenting the text instance and subsequently group the segments into a box through post-processing.

\noindent\textbf{Scene Text Spotting.} Scene text spotting represents the unification of detection and recognition processes within a singular framework. Numerous studies have improved performance by simultaneously learning detectors and recognizers. In \cite{liao2017textboxes,2018textboxes,2018rosetta,2018fots,2018end,2019textdragon}, end-to-end implementation is achieved by training the detection and recognition separately. The Mask TextSpotter \cite{2018masktextspotv1,2021masktextspotv2,2020masktextspotv3} series performs character segmentation during the recognition process. The ABCNet \cite{2020abcnet,2021abcnetv2} series obtains detection coordinates through control points of the Bezier curve. SwinTextSpotter \cite{2022swintextspotter} and ESTextSpotter \cite{2023estextspotter} use separate detection-recognition heads corresponding to different annotation formats. 

Furthermore, some methods directly decode coordinates and recognition results directly by predicting sequence, utilizing the structure of transformer encoder and decoder. Both SPTS \cite{2022spts} and SPTS V2 \cite{2023sptsv2} obtain coordinates by predicting the central point of the text instance, employing an auto-regressive approach to predict the central point and word transcription tokens. UNITS \cite{2023units} can handle various types of detection formats through prompts, and they can extract text beyond the number of trained text instances. DeepSolo \cite{2023deepsolo} inspired by DETR, enables the decoder to simultaneously perform text detection and recognition.

\noindent\textbf{Vision-Language Pre-training.} Modern pre-training methods generally involve MLM \cite{bert} or MIM \cite{mae, efficient,simmim}, or a combination of both. The MLM task randomly masks a set of text tokens from the input and reconstructs them based on the context around the masked tokens. MIM, on the other hand, randomly masks a percentage of image patches and predicts the RGB values of raw pixels. STKM \cite{2021sktm} learns text knowledge from datasets with image-level text annotations, the acquired text knowledge can subsequently be transferred to various text detectors. Inspired by CLIP \cite{2021clip}, VLPT \cite{song2022vision} adopts fine-grained cross-modality interaction to align unimodal embeddings for learning better representations of backbone via carefully designed pre-training tasks. oCLIP \cite{language} proposes an MLM-based vision-language pre-training method, which has achieved excellent performance in text detection and text spotting tasks. StrucTexTv2 \cite{2023structextv2} implements pre-training through MIM, where it randomly masks some text word regions in the input images and feeds them into the encoder.

\begin{figure}[ht]
  \centering
  \begin{subfigure}{0.24\linewidth}
    \includegraphics[width=1.\linewidth]{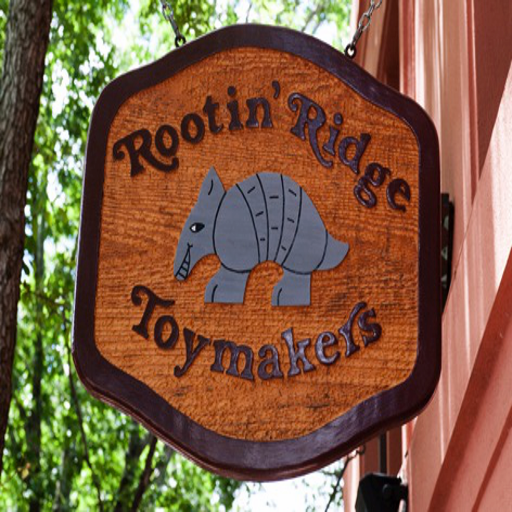}
    \caption{Origin}
    \label{fig:short-o}
  \end{subfigure}
  \begin{subfigure}{0.24\linewidth}
    \includegraphics[width=1.\linewidth]{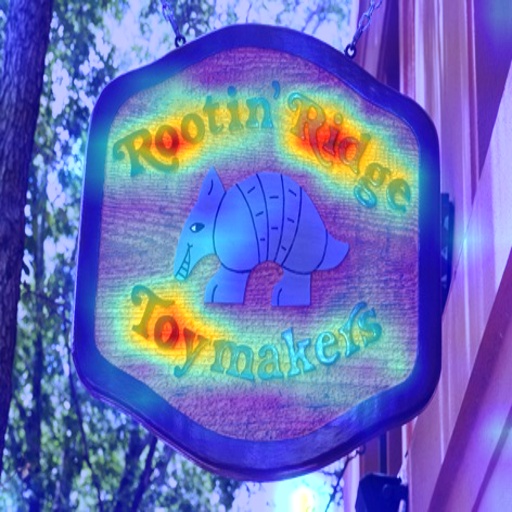}
    \caption{No change}
    \label{fig:short-a}
  \end{subfigure}
  \begin{subfigure}{0.24\linewidth}
    \includegraphics[width=1.\linewidth]{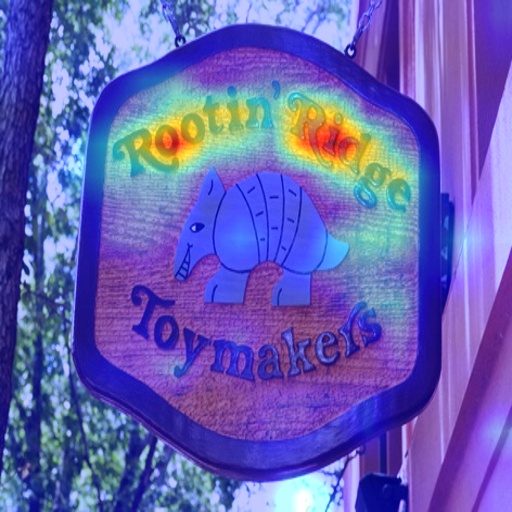}
    \caption{Drop-Text}
    \label{fig:short-b}
  \end{subfigure}
  \begin{subfigure}{0.24\linewidth}
    \includegraphics[width=1.\linewidth]{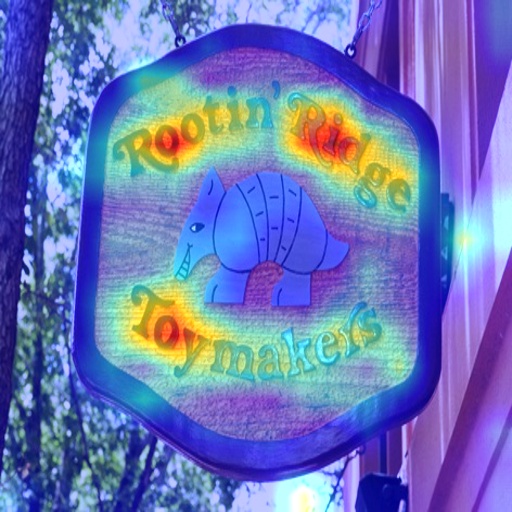}
    \caption{Noise-Text}
    \label{fig:short-c}
  \end{subfigure}
  \caption{Illustration of the proposed Text-Controller Module: The attention heatmap (from the cross-attention layer) of the text branch under different input scenarios is depicted. (a)The original image. (b) The text input consists of three instances: ``Rootin'', ``Ridge'', and ``Toymakers''. (c) The ``Toymakers'' instance is discarded. (d) A non-existent instance ``Sjehf'' is added.}
  \label{fig:textcontrol}
\end{figure}

While these methods have shown the effectiveness of pre-training in enhancing OCR performance, they either do not utilize higher-level textual semantic information as input or do not explicitly utilize the positional information of OCR-Text, making it difficult to achieve effective alignment between text and OCR-Text. Inspired by MaskFeat \cite{2022maskefeat}, which achieves pre-training by reconstructing the HOG features of masked image regions, and considering that glyph has been employed in some OCR tasks \cite{glyphdraw,textdiffuser,yang2024class,anytext} with proven efficacy. We propose utilizing the Text Controller module to reconstruct the corresponding destylized glyph of the text. This approach enables visible alignment between text and OCR-Text. As shown in \cref{fig:textcontrol}, our method can better attend to OCR-Text, highlighting its superiority in learning visual text representations for scene text image tasks.

\section{Methodology}

We introduce ODM, a pre-training technique that effectively aligns text and OCR-Text, leveraging the intrinsic characteristics of OCR-Text and explicitly learning the location information. The overall pipeline of our method is depicted in \cref{fig:pipeline}. Our network consists of two input branches: the image encoder, which employs ResNet50 \cite{2016resnet} to extract features from input images, and the text encoder, designed to extract textual features. By applying cross-attention between the extracted textual and visual features, we generate image features that align with textual features. These aligned features are then processed by a decoder to generate destylization binary images.

\subsection{The Text-Controller Module}
To address the challenge of the image encoder-decoder architecture in comprehending the significance of characters, we introduce the Text-Controller module to regulate the feature extraction of the image and align the OCR-Text features with textual features in the hidden space.

\noindent\textbf{Drop-Text.} General text encoders, such as CLIP \cite{2021clip}, are designed to process text descriptions as input, with the primary objective of establishing alignment between text and image features. In the Text-Controller Module, control over the decoder is executed through prompts.  Specifically, when a portion of the OCR-Text is input, the model is expected to only reconstruct that part of the binary image, treating the remaining OCR-Text as the background. During the training phase, the input OCR-Text is selected randomly, with a ratio varying from 0\% to 100\%. This strategy encourages the model to focus more on aligning the text with the corresponding OCR-Text.

\noindent\textbf{Noise-Text.} Noise, such as inaccurate labels, can potentially impact the model's training effectiveness. In contrast, we have introduced a concept termed Noise-Text, which leverages noise to augment the model's performance. This approach entails adding noise to the text encoder's input, introducing non-existent OCR-Text by altering its Token encoding value. The integration of these disruptive elements empowers the model to align the features of the text and OCR-Text more effectively, even in more intricate and challenging scenarios.

In \cref{fig:textcontrol}, we demonstrate the performance of cross-attention on images when employing different strategies in Text-Controller Module. Our approach achieves alignment between the text and corresponding OCR-Text while remaining unaffected by noisy text.

\subsection{OCR-Text Destylization}
To ensure that the image encoder can learn the fundamental features of OCR-Text, we have designed a simple decoder to reconstruct the destylized glyph of the text. This decoder comprises only basic FPN \cite{2017fpn} layer upsampling and 1*1 convolution. 

\begin{figure}[t]
  \centering
  \begin{subfigure}{0.32\linewidth}
    \includegraphics[width=1.\linewidth]{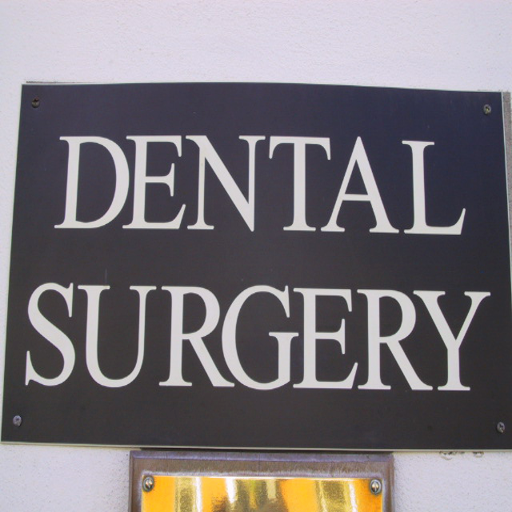}
    \label{fig:short-a}
  \end{subfigure}
  \begin{subfigure}{0.32\linewidth}
    \includegraphics[width=1.\linewidth]{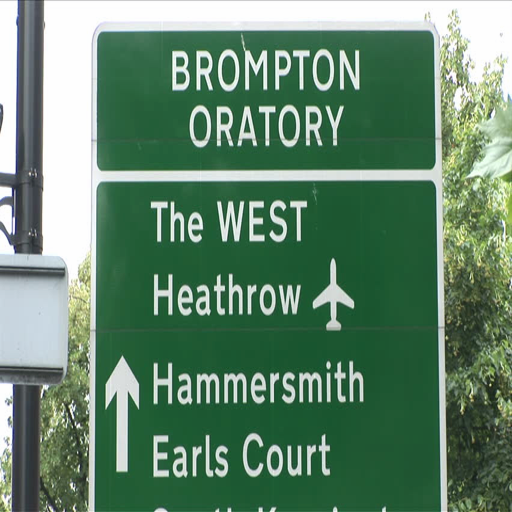}
    \label{fig:short-b}
  \end{subfigure}
  \begin{subfigure}{0.32\linewidth}
    \includegraphics[width=1.\linewidth]{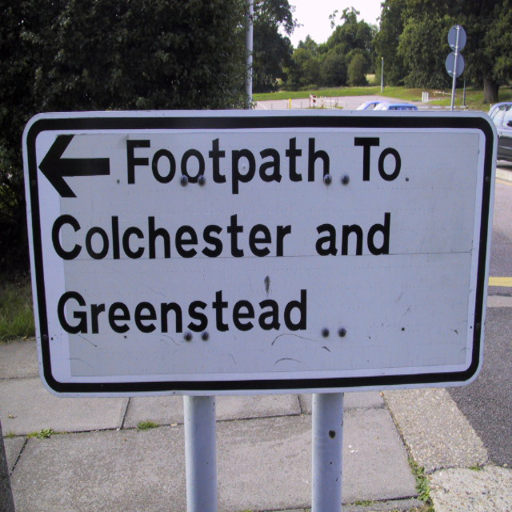}
    \label{fig:short-c}
  \end{subfigure}
  \begin{subfigure}{0.32\linewidth}
    \includegraphics[width=1.\linewidth]{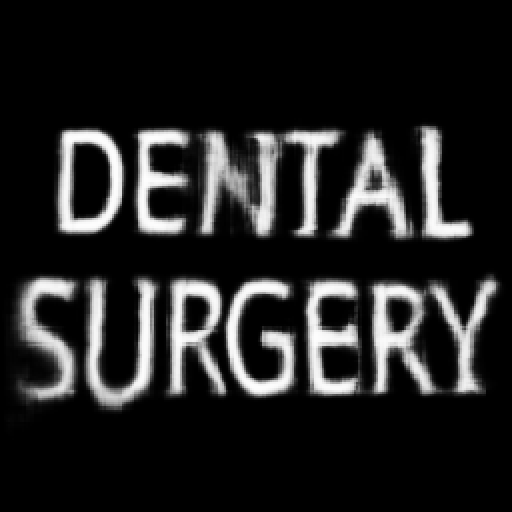}
    \label{fig:short-d}
  \end{subfigure}
  \begin{subfigure}{0.32\linewidth}
    \includegraphics[width=1.\linewidth]{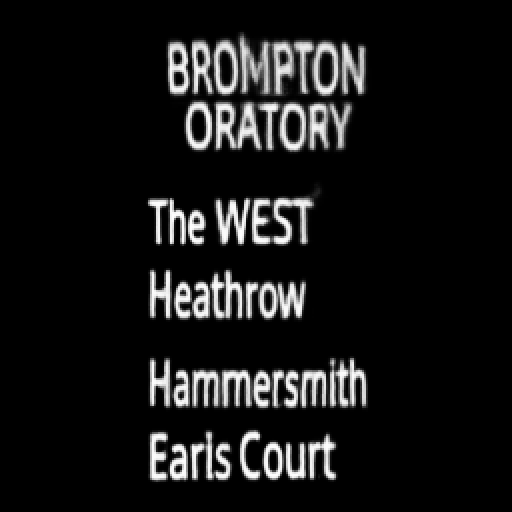}
    \label{fig:short-e}
  \end{subfigure}
  \begin{subfigure}{0.32\linewidth}
    \includegraphics[width=1.\linewidth]{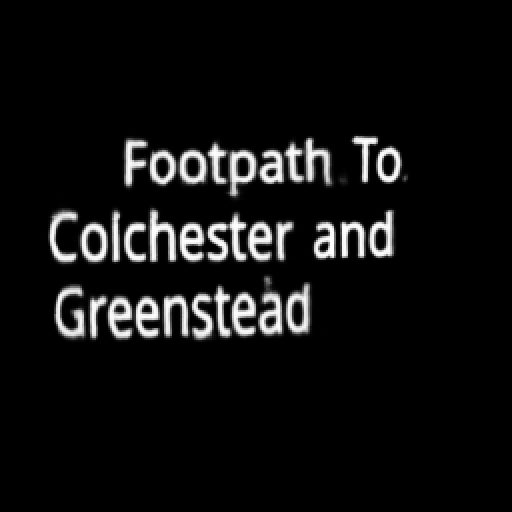}
    \label{fig:short-f}
  \end{subfigure}
  \caption{The upper row and lower row represent the original images and their corresponding predicted results, respectively. }
  \label{fig:pred}
\end{figure}

%
\cref{fig:pred} demonstrates the predicted results on some real images using our proposed method. The training dataset only consists of SynthText \cite{gupta2016synthetic} and does not include these real images. From the results, it can be observed that our proposed method effectively achieves OCR-Text destylization.

\subsection{Loss Function}
The ODM produces a binary image, conceptualizing supervised training as a pixel-level segmentation task. As a result, we optimize the model using a binary cross-entropy loss function for training:
\begin{equation}
{\mathcal{L}_{seg}}=\frac{1}{N}\sum_{i=1}^{N}-\left [ {y}_{i}\log({p}_{i}) + (1-{y}_{i})\log(1-{p}_{i})\right ]  
\label{eq:bce}
\end{equation}
where $N$ represents the number of pixels, $p_i$ represents the predicted result, and $y_i$ represents the Ground Truth.

On the other hand, the training output of ODM is to generate a destylization binary image, rather than a text segmentation map that maintains proportional consistency with the original image. Consequently, relying exclusively on pixel-level cross-entropy loss poses a challenge in effectively guiding the model to learn the destylization of characters. To optimize at the feature level, we incorporate the OCR LPIPS loss function proposed by OCR-VQGAN \cite{2023ocrgan}. Specifically, we utilize a well-trained detector (Unet-VGG16\cite{2014vgg16}) to input both the Ground Truth and the model's predicted binary images. Subsequently, the $\mathcal{L}$1 Loss is computed to foster the learning of a rich latent space and the destylized character glyph images, which is defined as follows:
\begin{equation}
{\mathcal{L}_{ocr}}=\sum_{l}\frac{1}{H_lW_l}\left \| {VGG}_{l}(\hat{y})-{VGG}_{l}(y) \right \|_{1}  
\label{eq:ocrloss}
\end{equation}
where $H_l$ and $W_l$ respectively represent the height and width of the output feature map in the l-th layer, $VGG$ represents a well-trained detector, $\hat{y}$ represents the predicted value, and $y$ represents the Ground Truth.

At the same time, drawing upon the batch-level contrastive loss proposed by CLIP, we expedite the mapping of text and images into the same semantic space by maximizing the similarity among positive samples and minimizing that among negative samples, which is defined as follows:
\begin{equation}
\begin{split}
{\mathcal{L}_{bc}}&=CE(\frac{exp(I,{T}_{+})}{\textstyle\sum_{i=1}^{B}exp(I,{T}_{i})},y(I)) \\
&\quad + CE(\frac{exp(T,{I}_{+})}{\textstyle\sum_{i=1}^{B}exp(T,{I}_{i})},y(T))
\end{split}
\label{eq:bc}
\end{equation}
where $CE$ represents the cross-entropy loss, $I$ represents image features, $T$ represents text features, $I_+$ represents the image feature corresponding to the current text feature, $T_+$ represents the text feature corresponding to the current image feature, and $y(*)$ represents the Ground Truth.

The loss function $\mathcal{L}$\textsubscript{total} is the sum of segmentation loss $\mathcal{L}$\textsubscript{seg}, OCR LPIPS loss $\mathcal{L}$\textsubscript{ocr}, and batch contrastive loss $\mathcal{L}$\textsubscript{bc}, formulated as follows:
\begin{equation}
\mathcal{L}_{total}=\alpha \mathcal{L}_{seg} + \beta \mathcal{L}_{ocr} + \gamma \mathcal{L}_{bc}
\label{eq:all}
\end{equation}
where the weights $\alpha$, $\beta$, and $\gamma$ are empirically set to 1, 1, and 0.5, respectively. 

\subsection{Label Generation}
One of the challenges in supervised training for ODM is the creation of data labels, particularly the acquisition of fine-grained pixel-level labels, which is both challenging and costly. To address this, we propose a method for generating pixel-level labels.

For four-point annotations, we calculate the dimensions of the quadrilateral and estimate the size and position of each character based on the number of characters. We then populate these characters using a font file, such as ``NotoSans-Regular". For multi-point annotations, we employ the image synthesis approach from ABCNet \cite{2020abcnet} and utilize the Bezier curves provided by the labels to compute the curvature of the text. Similar to four-point annotations, we determine the size and position of each character from the multi-point annotations and calculate the slope using the top-left point of each character and the subsequent point. Once we obtain the slope, we adjust the orientation of the characters accordingly. Some examples of the generated labels are shown in \cref{fig:label_gen}.

During the process of acquiring pixel-level labels, there may be discrepancies in the spacing between the original OCR-Text and the pixel-level labels. As a result, there might be instances where individual characters in our generated labels do not align precisely with the position of OCR-Text characters in the original image. However, this discrepancy does not impact our task since our approach involves transforming the original image into a destylization binary image rather than performing pixel-by-pixel text segmentation.

\section{Experiment}

\subsection{Datasets}
In our experiments, we employ a substantial quantity of publicly accessible datasets, executing pre-training on the SynthText \cite{gupta2016synthetic} dataset and fine-tuning on text detection and text spotting tasks utilizing datasets like ICDAR15 \cite{karatzas2015icdar}, CTW1500 \cite{liu2019curved}, and TotalText \cite{ch2020total}.

\noindent\textbf{SynthText} is a synthetic dataset for text detection and recognition, consisting of more than 800K synthesized images. Each image is annotated with text regions and their corresponding textual content. The synthetic texts in the dataset are generated by integrating real texts into natural images. The positions, orientations, sizes, and appearances of the synthetic texts are randomly varied to simulate text in real-world scenarios.

\noindent\textbf{ICDAR2015} consists of 1500 images, allocates 1000 images for training, and the remaining 500 for testing. The ICDAR2015 contains different types of text, including horizontal text and multi-oriented text.

\noindent\textbf{CTW1500} is a dataset for curved text instances, consisting of 1500 images, comprising 1500 images, of which 1000 are allocated for training and the remaining 500 for testing.

\noindent\textbf{TotalText} is a dataset for curved text instances, consisting of 1555 images, with 1255 designated for training and the remaining 300 for testing.


\begin{table*}[ht]
  \centering
  \caption{The performance of different models on ICDAR15, CTW1500, and TotalText for scene text detection is presented in the table. ``PD" and ``Syn" refer to the pre-training dataset and SynthText dataset, respectively. ``+ODM" refers to our pre-trained model on the SynthText dataset, which is adopted for fine-tuning. {$\dagger$} represents the scores obtained after reproducing the model, and $\varDelta$ represents the improvement.}
  \resizebox{1\textwidth}{!}{
    \begin{tabular}{ccccccccccc}
    \toprule
    \multirow{2}{*}{Method} &
    \multirow{2}{*}{PD} &
    \multicolumn{3}{c}{ICDAR 15} &
    \multicolumn{3}{c}{CTW1500} &
    \multicolumn{3}{c}{TotalText} \\
    \cmidrule(lr){3-5}
    \cmidrule(lr){6-8}
    \cmidrule(lr){9-11}
    & &
    \multicolumn{1}{c}{P} & \multicolumn{1}{c}{R} & \multicolumn{1}{c}{H} & 
    \multicolumn{1}{c}{P} & \multicolumn{1}{c}{R} & \multicolumn{1}{c}{H} & 
    \multicolumn{1}{c}{P} & \multicolumn{1}{c}{R} & \multicolumn{1}{c}{H} \\
    \midrule
    FCENet\cite{2021fce}{$\dagger$} & - & 82.43 & \textbf{88.34} & 85.28 & \textbf{85.7} & 80.1 & 83.1 & 78.1 & \textbf{84.85} & 81.34 \\
    FCENet + ODM & Syn & \textbf{91.38} & 82.19 & \textbf{86.54} & 84.92 & \textbf{82.33}  & \textbf{83.60} & \textbf{84.05} & 80.67 & \textbf{82.32} \\
    $\varDelta$ &  &  &  & \textcolor{red}{+ 1.26}& &  & \textcolor{red}{+ 0.5} &  &  & \textcolor{red}{+ 0.98} \\
    \midrule
    DBNet++\cite{2022dbnetpp}{$\dagger$} & - & 91.61 & 81.46 & 86.24 & 79.06 & 79.50 & 79.28 & 81.84 & 80.22 & 81.02 \\
    DBNet++ + ODM & Syn & \textbf{91.87} & \textbf{85.94} & \textbf{88.81} & \textbf{81.99} & \textbf{81.98}  & \textbf{81.97} & \textbf{88.01} & \textbf{82.25} & \textbf{85.03} \\
    $\varDelta$ &  &  &  & \textcolor{red}{+ 2.57}& &  & \textcolor{red}{+ 2.69} &  &  & \textcolor{red}{+ 4.01} \\
    \bottomrule              
    \end{tabular}
  }
  \label{tab:det}
\end{table*}

\begin{table*}
  \centering
  \caption{Comparison with existing scene text pre-training techniques on PSENet. ``PD" and ``Syn" refer to the pre-training dataset and SynthText dataset, respectively. ``+ODM" refers to our pre-trained model on the SynthText dataset, which is adopted for fine-tuning. {$\dagger$} represents the scores obtained after reproducing the model, and $\varDelta$ represents the improvement.}
  \resizebox{1\textwidth}{!}{
    \begin{tabular}{ccccccccccc}
    \toprule
    \multirow{2}{*}{Method} &
    \multirow{2}{*}{PD} &
    \multicolumn{3}{c}{ICDAR 15} &
    \multicolumn{3}{c}{CTW1500} &
    \multicolumn{3}{c}{TotalText} \\
    \cmidrule(lr){3-5}
    \cmidrule(lr){6-8}
    \cmidrule(lr){9-11}
    & &
    \multicolumn{1}{c}{P} & \multicolumn{1}{c}{R} & \multicolumn{1}{c}{H} & 
    \multicolumn{1}{c}{P} & \multicolumn{1}{c}{R} & \multicolumn{1}{c}{H} & 
    \multicolumn{1}{c}{P} & \multicolumn{1}{c}{R} & \multicolumn{1}{c}{H} \\
    \midrule
    PSENet\cite{2019pse}{$\dagger$} & - & 83.96 & 76.36 & 79.98 & 80.6 & 75.6 & 78.0 & 75.1 & 81.8 & 78.3 \\
    PSENet\cite{2019pse} & Syn & 86.2 & 79.4 & 82.7 & 81.8 & 77.8 & 79.7 & 87.8 & 79.0 & 82.6 \\
    PSENet + STKM\cite{2021sktm} & Syn & 87.78 & 84.06 & 85.88 & 85.08 & 78.23 & 81.51 & 85.08 & 78.23 & 81.51 \\
    PSENet + oCLIP\cite{language} & Syn & \textbf{88.95} & 80.98 & 84.78 & 86.3 & 79.6  & 82.8 & 90.7 & 80.8 & 85.5 \\
    PSENet + oCLIP\cite{language} & Web & - & - & - & \textbf{87.5} & 79.9 & 83.5 & \textbf{92.2} & 82.4 & \textbf{87.0} \\
    \midrule
    PSENet + ODM & Syn & 88.43 & \textbf{85.41} & \textbf{86.90} & 85.86 & \textbf{85.38} & \textbf{85.62} & 88.56 & \textbf{83.37} & 85.94 \\
    $\varDelta$ &  &  &  & \textcolor{red}{+ 6.92}& &  & \textcolor{red}{+ 7.62} &  &  & \textcolor{red}{+ 7.64} \\

    \bottomrule              
    \end{tabular}
  }
  \label{tab:pre}
\end{table*}

\subsection{Implementation Details}
\noindent\textbf{Training.} In our proposed method, we employ ResNet50 as the image encoder, with a 6-layer transformer \cite{2017transformer} serving as the text encoder. The reconstruction decoder is constructed from FPN and 1x1 convolutions. During the training phase, we configure the image size to 512x512, set the text length for the Text-Controller to 25, and cap the maximum number of text instances at 32. The learning rate is established at 1e-4. The network training is conducted on 8 A100 GPUs, with each card handling a batch size of 64, resulting in a total batch size of 512. The entire training process spans 100 epochs.

\noindent\textbf{Fine-tuning.} We executed evaluations on a variety of OCR tasks, encompassing text detection methods such as DB++ \cite{2022dbnetpp}\footnote{\url{https://github.com/open-mmlab/mmocr/tree/main/configs/textdet/dbnetpp}}, FCENet \cite{2021fce}\footnote{\url{https://github.com/open-mmlab/mmocr/tree/main/configs/textdet/fcenet}}, and PSENet \cite{2019pse}\footnote{\url{https://github.com/open-mmlab/mmocr/tree/main/configs/textdet/psenet}}, as well as text spotting techniques including ABCNet \cite{2020abcnet}\footnote{\url{https://github.com/aim-uofa/AdelaiDet/blob/master/configs/BAText}}, DeepSolo \cite{2023deepsolo}\footnote{\url{https://github.com/ViTAE-Transformer/DeepSolo}}, and SPTS \cite{2022spts}\footnote{\url{https://github.com/shannanyinxiang/SPTS}}. 
In the comparative experiments, the only variable altered is the backbone weights, with all other settings maintained consistently.

\noindent\textbf{Evaluation protocol}. We use intersection over union (IoU) to determine whether the model correctly detects the region of text, and we calculate precision (P), recall (R), and Hmean (H) for comparison following ICDAR2015 \cite{karatzas2015icdar}.

\subsection{Comparison with Detection Methods}
To evaluate the effectiveness of our proposed method on text detection tasks, we conducted comparable experiments with DB++, PSENet, and FCENet. Our model underwent initial pre-training on the SynthText dataset and was subsequently fine-tuned on different datasets. The results of these experiments are shown in \cref{tab:det} and \cref{tab:pre}. The networks utilizing our pre-trained backbone demonstrated improvements compared to their original counterparts, with PSENet showing particularly notable improvement. This suggests that our pre-trained model weights can effectively focus on text regions within the scene, showcasing the significant potential of our pre-trained weights.

\begin{table}
  \centering
  \caption{The performance of different models on ICDAR15 for scene text spotting is presented in the table. ``PD" and ``Syn" refer to the pre-training dataset and SynthText dataset, respectively. ``+ODM" refers to our pre-trained model on the SynthText dataset, which is adopted for fine-tuning. {$\dagger$} represents the scores obtained after reproducing the model, and $\varDelta$ represents the improvement. `D' and `E' represent Detection and End-to-End, respectively. `S', `W', and `G' donate using strong, weak, and generic lexicons, respectively.}
  \resizebox{.47\textwidth}{!}{
    \begin{tabular}{cccccc}
    \toprule
    \multirow{1}{*}{Method} &
    \multirow{1}{*}{PD} &
    \multicolumn{1}{c}{D-H} & \multicolumn{1}{c}{E-S} & \multicolumn{1}{c}{E-W} & \multicolumn{1}{c}{E-G} \\
    \midrule
    ABCNet\cite{2020abcnet}{$\dagger$} & - & 85.22 & 78.01 & 73.86 & 68.09 \\
    ABCNet + ODM & Syn & \textbf{87.56} & \textbf{80.87} & \textbf{75.75} & \textbf{70.01}  \\
    $\varDelta$ &  & \textcolor{red}{+ 2.34} & \textcolor{red}{+ 2.86} & \textcolor{red}{+ 1.89} & \textcolor{red}{+ 1.92} \\

    \midrule
    DeepSolo\cite{2023deepsolo}{$\dagger$} & - & 86.44 & 83.50 & 79.36 & 74.19 \\
    DeeoSolo + ODM & Syn & \textbf{88.08} & \textbf{85.83} & \textbf{80.67} & \textbf{75.2}  \\
    $\varDelta$ &  & \textcolor{red}{+ 1.64} & \textcolor{red}{+ 2.3} & \textcolor{red}{+ 1.31} & \textcolor{red}{+ 1.01} \\

    \midrule
    SPTS\cite{2022spts}{$\dagger$} & - & - & 77.5 & 70.2 & 65.8  \\
    SPTS + ODM & Syn & - & \textbf{78.8} & \textbf{72.1} & \textbf{67.8} \\
    $\varDelta$ & & & \textcolor{red}{+ 1.3} & \textcolor{red}{+ 1.9} & \textcolor{red}{+ 2.0} \\

    \bottomrule              
    \end{tabular}
  }
  \label{tab:rec-ic15}
\end{table}

\begin{table}
  \centering
  \caption{The performance of different models on CTW1500 for scene text spotting is presented in the table. ``PD" and ``Syn" refer to the pre-training dataset and SynthText dataset, respectively. ``+ODM" refers to our pre-trained model on the SynthText dataset, which is adopted for fine-tuning. {$\dagger$} represents the scores obtained after reproducing the model, and $\varDelta$ represents the improvement. `D' and `E' represent Detection and End-to-End, respectively. `N' and `F' represent None and Full, respectively.}
  \resizebox{.47\textwidth}{!}{
    \begin{tabular}{cccccc}
    \toprule
    \multirow{1}{*}{Method} &
    \multirow{1}{*}{PD} &
    \multicolumn{1}{c}{D-H} & \multicolumn{1}{c}{E-N} & \multicolumn{1}{c}{E-F} \\
    \midrule
    ABCNet\cite{2020abcnet}{$\dagger$} & - & 83.69 & 64.17 & 78.66 \\
    ABCNet + ODM & Syn & \textbf{85.40} & \textbf{65.06} & \textbf{79.79} \\
    $\varDelta$ &  & \textcolor{red}{+ 1.71} & \textcolor{red}{+ 0.89} & \textcolor{red}{+ 1.13} \\

    \midrule
    DeepSolo\cite{2023deepsolo}{$\dagger$} & - & 86.31 & 76.35 & 84.31 \\
    DeeoSolo + ODM & Syn & \textbf{86.58} & \textbf{78.07} & \textbf{85.65} \\
    $\varDelta$ &  & \textcolor{red}{+ 0.27} & \textcolor{red}{+ 1.72} & \textcolor{red}{+ 1.34} \\

    \midrule
    SPTS\cite{2022spts}{$\dagger$} & - & - & 74.2 & 82.4 \\
    SPTS + ODM & Syn & - & \textbf{78.2} & \textbf{84.2} \\
    $\varDelta$ &  &  & \textcolor{red}{+ 4.0} & \textcolor{red}{+ 1.8} \\

    \bottomrule              
    \end{tabular}
  }
  \label{tab:rec-totaltext}
\end{table}
\subsection{Comparison with Spotting Methods}
To evaluate the effectiveness of our proposed method on text spotting tasks, we conducted comparable experiments with ABCNet, DeepSolo, and SPTS, covering a wide range of task scenarios. Our model was initially pre-trained on the SynthText dataset and then fine-tuned on different datasets. The results of these experiments are presented in \cref{tab:rec-ic15} and \cref{tab:rec-totaltext}. The experimental results demonstrate that utilizing our pre-trained weights for fine-tuning significantly improves the performance of the models across various datasets. This indicates that the pre-trained weights acquired through our proposed pre-training method accurately locate the positions of OCR-Text and extract image features that capture the semantic information of the text instance.

\subsection{Weakly Supervised Pre-training}
To evaluate the effectiveness of our proposed method in aligning text with images in weakly labeled data, we assess its efficacy. We first employ PPOCRv3 \cite{li2022pp} to generate pseudo labels (i.e., text and position in the image) on the 400,000 weakly annotated images from the LSVT \cite{sun2019icdar} dataset. We then generate the corresponding pixel-level destylized annotations using our proposed label generation method. To ensure the quality of pre-training, we only select labels with inference confidence exceeding 0.9 and a text size larger than 32 pixels. We pre-train our model with these generated labels and subsequently fine-tune different scene text detectors using 30,000 fully annotated images from the LSVT dataset. The results presented in \cref{tab:weak} demonstrate that our proposed method, when exclusively utilizing the pseudo labels for pre-training, achieves approximately a 1.5\% improvement in the Hmean score after fine-tuning on both DBNet++ and PSENet models. This demonstrates that our method can perform consistently even under weakly supervised scenarios and partially addresses the issue of a large amount of unlabeled data not being able to participate in pre-training.

\begin{table}
  \centering
  \caption{The performance of different models on LSVT for scene text detection is presented in the table. ``PD" and ``LSVT" refer to the pre-training dataset and LSVT dataset, respectively. ``+ODM" refers to our pre-trained model with 400,000 pseudo-label images in the LSVT dataset, which is adopted for fine-tuning. {$\dagger$} represents the scores obtained after reproducing the model, and $\varDelta$ represents the improvement.}
  \resizebox{.47\textwidth}{!}{
    \begin{tabular}{ccccc}
    \toprule
    \multirow{1}{*}{Method} & \multirow{1}{*}{PD} &\multirow{1}{*}{P} & \multirow{1}{*}{R} & \multirow{1}{*}{H} \\
    \midrule
    PSENet\cite{2019pse}{$\dagger$} & - & 64.61 & 72.29 & 68.23 \\
    PSENet + ODM & LSVT & \textbf{66.33} & \textbf{73.60} & \textbf{69.78} \\
    $\varDelta$ &  &  &  & \textcolor{red}{+ 1.55} \\

    \midrule
    DBNet++\cite{2022dbnetpp}{$\dagger$} & - & 71.21 & \textbf{79.28} & 75.02  \\
    DBNet++ + ODM & LSVT & \textbf{75.81} & 77.16 & \textbf{76.48}  \\   
    $\varDelta$ &  &  &  & \textcolor{red}{+ 1.46} \\

    \bottomrule              
    \end{tabular}
  }
  \label{tab:weak}
\end{table}
\subsection{Comparison with Pre-training Methods}
We compare our proposed method with existing scene text pre-training strategies, including STKM and oCLIP. To investigate the effectiveness of different pre-training objectives, we conducted ablative experiments with PSENet on three datasets: ICDAR15, CTW1500, and TotalText. As shown in \cref{tab:pre}, when pre-training on the same set of data, our proposed method outperforms the existing pre-training techniques. 
Furthermore, when fine-tuning on the CTW1500 dataset, our proposed method, which was pre-trained on SynthText alone, even surpasses the performance of oCLIP, which was pre-trained on 40 million web images.


\begin{table}
  \centering
  \caption{Proportion ablation study of the proposed Text-Controller module on CTW1500. ``PD" and ``Syn" refer to the pre-training dataset and SynthText dataset, respectively. ``TP" refers to the selected text proportion. ``+ODM" refers to our pre-trained model on the SynthText dataset, which is adopted for fine-tuning. {$\dagger$} represents the scores obtained after reproducing the model.}
  \resizebox{.47\textwidth}{!}{
    \begin{tabular}{cccccc}
    \toprule
    Method & PD & TP & P & R & H \\
    \midrule
    PSENet\cite{2019pse} & - & - & 80.6 & 75.6 & 78.0 \\
    PSENet\cite{2019pse} & Syn & - & 81.8 & 77.8 & 79.7 \\
    \midrule
    PSENet + ODM & Syn & 0\% & 82.41 & 84.19 & 83.29  \\
    PSENet + ODM & Syn & 30\% & 84.83 & 82.18 & 83.48  \\
    PSENet + ODM & Syn & 50\% & 83.76 & \textbf{84.78} & \textbf{84.27}  \\
    PSENet + ODM & Syn & 70\% & \textbf{85.32} & 82.37 & 83.82  \\
    \bottomrule              
    \end{tabular}
  }
  \label{tab:ab1}
\end{table}

\begin{table}
  \centering
  \caption{Ablation study of our proposed components on TotalText. We fine-tune PSENet by using the pre-trained models with different modules. ``TE", ``DT", ``NT", and ``OL" refer to Text Encoder, Drop-Text, Noise-Text, and OCR Loss, respectively.}
  \resizebox{.47\textwidth}{!}{
    \begin{tabular}{cccccccc}
    \toprule
    Method & TE & DT & NT & OL & P & R & H \\
    \midrule
    PSENet\cite{2019pse} &  &  &  &  & 75.10 & 81.80 & 78.30 \\
    \midrule
    PSENet + ODM &  &  &  &  & 85.08 & 78.55 & 81.68  \\
    PSENet + ODM & {\checkmark} &  &  &  & 86.66 & 82.75 & 84.66  \\
    PSENet + ODM & {\checkmark} & {\checkmark} &  &  & 88.06 & 82.97 & 85.44  \\
    PSENet + ODM & {\checkmark} &  & {\checkmark} &  & 88.10 & 82.57 & 85.24  \\
    PSENet + ODM & {\checkmark} & {\checkmark} & {\checkmark} &  & 88.17 & 83.21 & 85.62  \\
    PSENet + ODM & {\checkmark} & {\checkmark} & {\checkmark} & {\checkmark} & \textbf{88.56} & \textbf{83.37} & \textbf{85.94}  \\
    \bottomrule              
    \end{tabular}
  }
  \label{tab:ab2}
\end{table}

\subsection{Ablation Experiments}
\textbf{Proportion Ablation} We conducted experiments to assess the influence of selected proportions in our Drop-Text and Noise-Text strategies. Four groups of experiments were performed with selected proportions of 0\%, 30\%, 50\%, and 70\% respectively. The selected proportions for these strategies were the only variables adjusted, while all other configurations remained constant. We pre-trained the models on the SynthText dataset with varying proportions and transferred the pre-trained weights to fine-tune PSENet on the CTW1500 dataset. The results are presented in \cref{tab:ab1}. The performance of the model was effectively enhanced by facilitating text-image alignment through the Drop-Text and Noise-Text strategies. The impact of varying proportions on performance was substantial. A small proportion of text instances resulted in minimal performance improvement, possibly due to the insufficient changes in a small number of words to achieve effective text-image alignment. This minimal improvement can be attributed to the errors introduced during training due to the lack of significant changes in the text instances. On the other hand, using a large proportion of text instances also had a limited effect on performance. This can be attributed to the fact that when a substantial number of words are either eliminated or added as noise, the model captures fewer valid features during training, leading to difficulties in convergence and potentially biased learning. Hence, it is important to select an appropriate proportion that ensures the model captures an adequate number of valid features. This helps the model align text with OCR-Text in more complex scenarios and enhances the robustness of the model.

\noindent\textbf{Module Ablation:} In our research, we investigated the contributions of several proposed modules. We trained the models with different combinations of these modules on the SynthText dataset and subsequently fine-tuned the pre-trained weights using PSENet on the TotalText dataset, as shown in \cref{tab:ab2}. The empirical results indicated that for the task of OCR-Text Desytlization, the text feature furnished by the Text-Controller module aids the model in better understanding and locating OCR-Text. Additionally, our proposed Drop-Text and Noise-Text strategies effectively bolster the model's performance by intensifying the alignment between text and OCR-Text.

\section{Conclusion}


This paper introduces ODM, a novel pre-trained method that aims to transform diverse styles of text found in images into a uniform style based on the text prompt. By leveraging the pre-trained model generated through ODM, it can seamlessly integrate into existing detection and spotting networks, resulting in significant performance improvements. To address the challenge of annotation costs in OCR tasks, we propose a new labeling generation method designed specifically for ODM. Additionally, we introduce the Text-Controller module, which helps regulate the model output and improves its understanding of OCR-Text. By combining these approaches, we enable a larger amount of unlabeled data to be used in the pre-training process, effectively reducing annotation costs. Our extensive ablation and comparative experiments demonstrate the effectiveness and robustness of our model. The results highlight the potential of ODM in OCR pre-training and its valuable contributions to the advancement of scene text detection and spotting tasks. Looking ahead, we plan to explore the potential of this method in other domains, such as document analysis, handwriting recognition, and other complex scene text scenarios. 
{
    \small
    \bibliographystyle{ieeenat_fullname}
    \bibliography{main}
}


\end{document}